\pdfoutput=1
\documentclass[11pt]{article}
\usepackage[table]{xcolor}
\usepackage[preprint]{acl}

\usepackage{times}
\usepackage{latexsym}
\usepackage{tikz}
\usepackage[T1]{fontenc}
\usepackage[utf8]{inputenc}
\usepackage{amsmath}
\usepackage{amssymb}
\usepackage{subcaption}
\usepackage{microtype}

\usepackage{inconsolata}
\usepackage{multirow}

\usepackage{graphicx}
\usepackage{booktabs}
\usepackage{listings}
\usepackage{makecell}

\usepackage{subcaption}     % <- provides \subcaptionbox

\newcommand{\datasetname}{\textsc{Disjoint-3DQA}}
\title{Out of Sight, Not Out of Context? \\
Egocentric Spatial Reasoning in VLMs Across Disjoint Frames}
\author{Sahithya Ravi$^{1}$\thanks{Work done during internship at MSR.}~~Gabriel Sarch $^{2} $~~Vibhav Vineet$^{3}$ \\ ~~\textbf{Andrew D. Wilson}$^{3}  $~~\textbf{Balasaravanan Thoravi Kumaravel}$^{3}$ \\
$^1$ University of British Columbia \quad $^2$ Carnegie Mellon University\\$^3$ Microsoft Research, Redmond\,WA\\
{\tt \small{sahiravi@cs.ubc.ca, gsarch@andrew.cmu.edu}}\\
{\tt \small\{vivineet, awilson, bala.kumaravel\}@microsoft.com}
}

\usepackage{enumitem}
\usepackage{tikz}

\usepackage{makecell}

\begin{document}
\maketitle

\begin{abstract}
An embodied AI assistant operating on egocentric video must integrate spatial cues across time -- for instance, determining where an \texttt{object A}, glimpsed  a few moments ago lies relative to an \texttt{object B} encountered later. We introduce \datasetname{}, a generative QA benchmark that evaluates this ability of VLMs by posing questions about object pairs that are not co-visible in the same frame. We evaluated seven state-of-the-art VLMs and found that models lag behind human performance by 28\%, with steeper declines in accuracy  (60\% $\rightarrow$ 30 \%) as the temporal gap widens. Our analysis further reveals that providing trajectories or bird's-eye-view projections to VLMs results in only marginal improvements, whereas providing oracle 3D coordinates leads to a substantial 20\% performance increase. This highlights a core bottleneck of multi-frame VLMs in constructing and maintaining 3D scene representations over time from visual signals. \datasetname{} therefore sets a clear, measurable challenge for long-horizon spatial reasoning and aims to catalyze
future research at the intersection of vision, language, and embodied AI.
\end{abstract}
% Methodology to generate data that can be scaled => ASE is scalable, as its synthetic, scalable in terms of QA pairs - pipeline 
% Permanence
% Distance graph
% Textual cues

% object permanence vs disjoint frame?
% mention that its scalable
% Sparse 3d vs full 3d - Asymmetry between 3D Text vs BEV images - augmentation of LLM

\section{Introduction}
\label{sec:Intro}

% edited 
We live in a three-dimensional world, and both humans and animals excel at building internal spatial representations that help them perceive, understand, and interact with their environments \cite{Wang2002HumanSR}. For machines to act as capable embodied assistants, they too must be able to reason spatially: to infer where objects are, how they relate to one another, and how to navigate through space \cite{cheng2024spatialrgpt, chen2024spatialvlm, cho2023spatially}. This is especially challenging in egocentric settings, where perception is anchored to a moving first-person viewpoint.

\begin{figure}[t!]
  \includegraphics[width=\columnwidth]{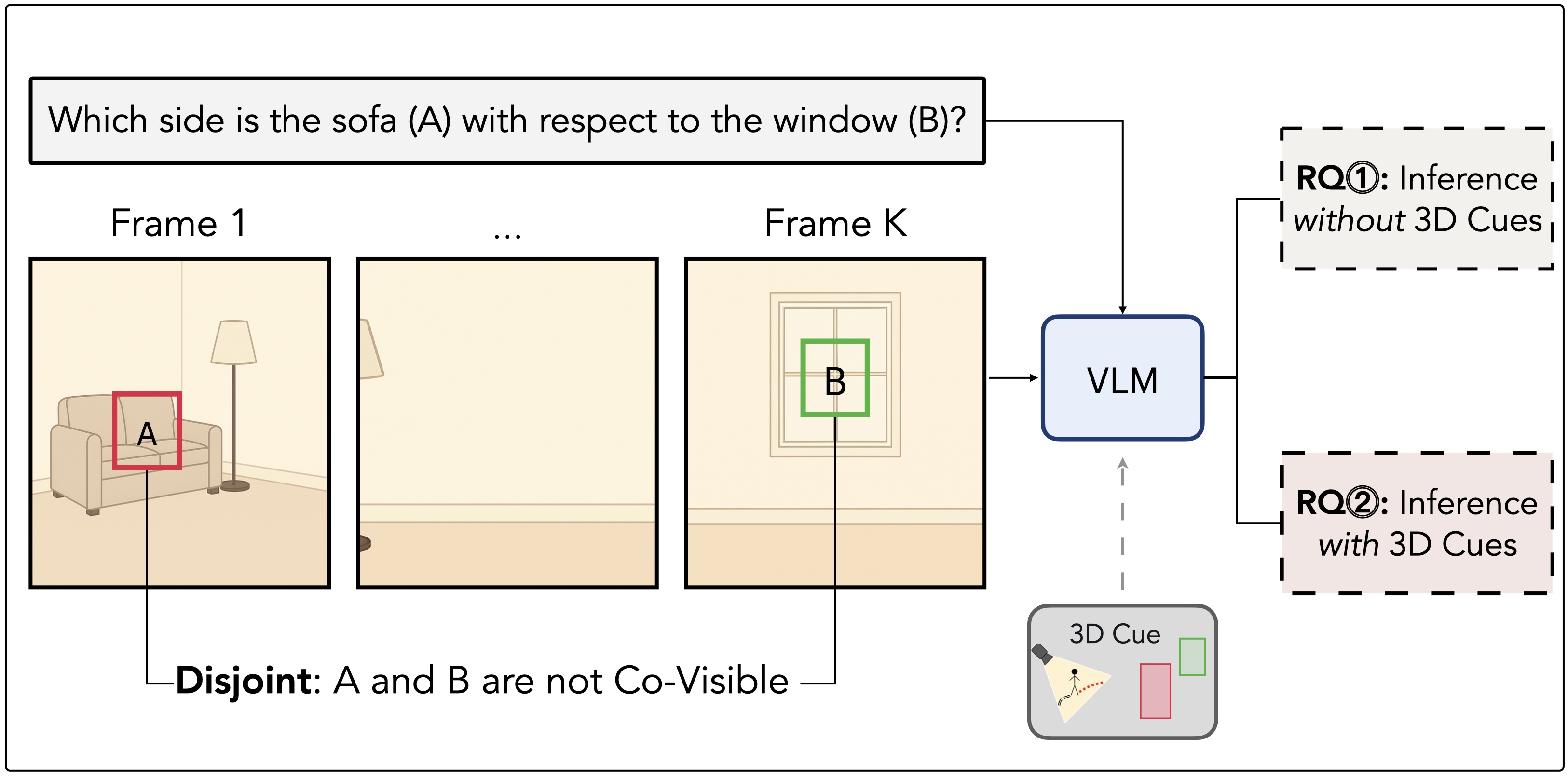}
  \caption{\datasetname{}: We focus on answering questions about spatial relationships when the objects are static but the camera is moving. }
  \label{fig:disjoint_concept}
\end{figure}
\begin{figure*}
  \includegraphics[width=0.95\textwidth]{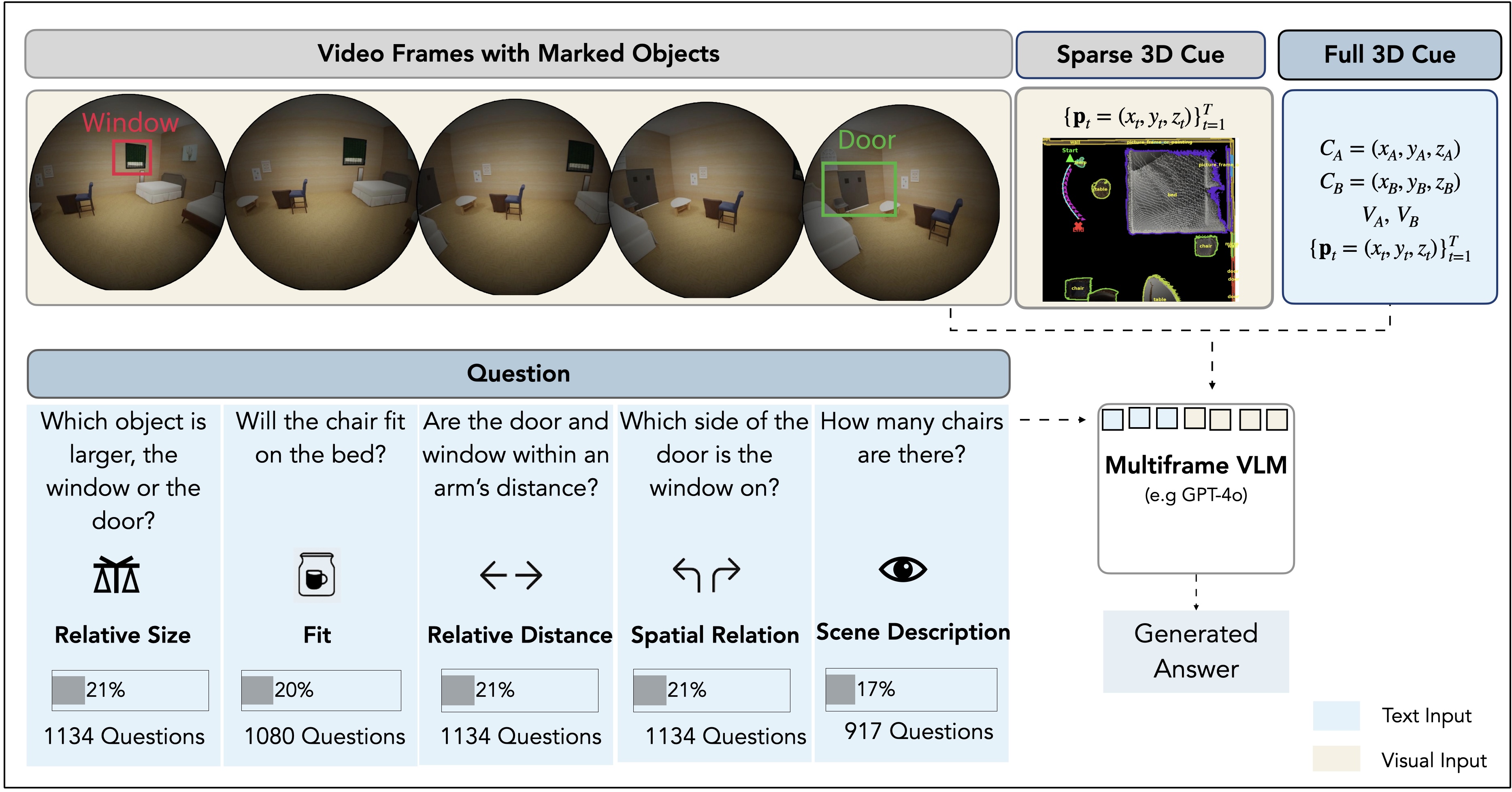}
  \caption{Demonstration of the evaluation setup of \datasetname{}. We provide the video frames with objects relevant to the question marked for visual grounding. Optionally, we evaluate models by providing explicit 3D cues using text or visual (Bird Eye View) inputs relevant to that question.}
  \label{fig:teaser}
\end{figure*}
As the camera wearer moves, objects may enter and exit the field of view at different times, requiring models to reason across \emph{temporally disjoint} observations. We term this setting \textbf{disjoint‑frame spatial reasoning}: a model must accumulate geometric cues across time, `mentally' reconstruct the scene, and then integrate across the cues to answer questions.  While related to object permanence tracking \cite{tokmakov2021learningtrackobjectpermanence}, our setting focuses on a harder subproblem, where objects do not co-occur, making spatial reasoning more challenging.

Figure~\ref{fig:disjoint_concept} illustrates a typical example: the camera wearer first views a sofa (object A) in one part of the room and only much later encounters a window (object B) from a different viewpoint. The question \textit{“Which side is the sofa (A) with respect to the window (B)?”} requires reasoning over temporally disjoint frames, where the two objects are not co-visible. This setup poses a fundamental challenge to current VLMs, which must infer spatial relationship among multiple frames. The core question we investigate is thus:  \textit{Can VLMs track and reason about spatial relationships when the relevant objects may not be co-visible?}

Recent work on embodied video understanding \cite{qiu-etal-2024-embodied, amin-rayz-2024-embodied, suglia-etal-2024-alanavlm, majumdar2024openeqa, cheng2024spatialrgpt} has shown that language-based reasoning alone is insufficient for improving performance, even when objects are co-visible. The challenge becomes more acute in 3D settings: without explicit spatial priors, VLMs must infer scene geometry from raw pixels, often producing brittle or hallucinated maps of the scene \cite{yang2024think}.

We introduce \datasetname{}\footnote{Dataset and code will be released upon publication.}, a generative QA benchmark that (i) poses spatial queries where relevant objects are not co-visible, compelling models to integrate information across frames and viewpoints and (ii) probe models with varying degree of explicit 3D scene information.  Constructed from RGB‑D egocentric recordings, \datasetname{} comprises 5,399 question–answer pairs spanning object-object relative direction, containment, and volumetric comparison, each requiring multi‑frame spatial integration. We evaluate a spectrum of proprietary and open‑source multi-frame VLMs including GPT‑4o and, Qwen\-2.5\-VL and observe an overall lag behind human performance by ~28\%. 
% At test time models receive only the RGB frames: video cameras are cheap and ubiquitous, whereas depth sensors remain costly, power-hungry, and scarce in training data

A natural hypothesis is that these models struggle because raw RGB frames provide limited geometric information, particularly in egocentric videos where relevant objects may never appear in the same frame due to continuous camera motion. We therefore augment the input with \textbf{explicit 3D context} in two forms: (i) \textit{Text Trajectory Cue}, through textual camera trajectories, and (ii) \textit{Top-down cue}, through bird’s-eye view (BEV) renderings that depict the camera’s path from object A to B. We refer to these as \textbf{sparse 3D cues}, because they expose only partial glimpses of the 3D scene geometry. While such cues lead to modest performance gains up to 2–3\%, we observe significantly larger improvements exceeding 20\%,  when models are provided with \textbf{full 3D context}, including ground-truth object coordinates, volumes, and spatial metadata.

This contrast points to a central open challenge: enabling models to treat spatial cues not as isolated tokens, but as grounded elements of a coherent metric space. \datasetname{} thus motivates the need for models that can build and maintain internal 3D representations from sparse, temporally disjoint egocentric inputs, a capacity critical for robust spatial understanding in real-world environments.

\section{\datasetname{}}
\label{sec:dataset}
Understanding grounded language begins with situating objects relative to one another, and to an embodied observer, within a coherent 3D world. Spatial reasoning for embodied agents typically arises in three settings: (i) when both camera and objects are static, (ii) when the camera moves through a static environment, and (iii) when both camera and objects move. We focus on the second setting, common in egocentric video, where a moving agent observes a static scene from varying perspectives.

To evaluate spatial reasoning in such scenarios, we introduce \datasetname{}, a benchmark designed to test whether models can integrate information across temporally disjoint frames where relevant objects never appear together. This setup reflects real-world contexts like AR navigation or assistive robotics, where a complete spatial view is rarely available at once. 

Figure~\ref{fig:teaser} shows an example: a model must infer the spatial relation between a window seen early and a door seen much later. Such questions require integrating partial cues over time to recover spatial structure. \datasetname{} contains 5,399 question–answer pairs across 1,668 scenes and 856 unique object pairs, with an average of four diverse questions per scene. Motivated by foundational work in spatial cognition \cite{Landau1993WhatA}, our questions span a range of reasoning types essential for embodied agents: relative direction, distance, size comparison, fit, and holistic scene understanding. Full dataset statistics are provided in Appendix~\ref{sec:dataset-stats}.

We build \datasetname{} using the Aria Synthetic Environments (ASE) dataset\footnote{\url{https://www.projectaria.com/datasets/ase/}}, a large-scale simulation of over 100,000 photorealistic indoor scenes out of which we randomly sample 1688 scenes. ASE offers rich 3D geometry, sensor metadata, and realistic object placements, making it ideal for spatial supervision. Its combination of realism, controllability, and scale enables high-precision question generation. Our construction pipeline is inherently \textbf{scalable} for two reasons: (a) it leverages the Aria Synthetic Environments, enabling controlled and repeatable data generation at scale, and (b) the steps in our pipeline are fully modular and parallelizable. 

\paragraph{Goal.}
Given a pair of objects $(A, B)$, the model must determine their spatial relation (e.g., left/right, size, relative distance) using only egocentric observations in which the objects appear separately. To construct such object pairs, we represent each video as a sequence of frames $\mathcal{F} = \{f_1, f_2, \dots, f_T\}$, where each frame $f_t$ contains a set of visible objects $\mathcal{O}_t$. For any object pair $(A, B)$, we define their visibility spans as:
\[
\mathcal{T}_A = \{ t \mid A \in \mathcal{O}_t \}, \quad
\mathcal{T}_B = \{ t \mid B \in \mathcal{O}_t \}
\]
We include $(A, B)$ in the dataset only if $\mathcal{T}_A \cap \mathcal{T}_B = \emptyset$, ensuring that the two objects are never seen in the same frame i.e. not \textit{co-visible}. This disjointness constraint forces models to accumulate spatial cues across non-overlapping views.
\paragraph{Ground-Truth Computation.}
To define ground-truth relations, we compute the position of object \( A \) relative to object \( B \) from the viewpoint of a frame where \( B \) is visible. Let \( \mathbf{T}_B \) be the world-to-camera transform for that frame, and let \( \mathbf{c}_A, \mathbf{c}_B \in \mathbb{R}^3 \) denote the objects' centers in world coordinates. We transform them into the camera coordinate frame as follows: \( \tilde{\mathbf{c}}_A = \mathbf{T}_B \cdot \mathbf{c}_A \) (object \( A \) in \( B \)'s frame), \( \tilde{\mathbf{c}}_B = \mathbf{T}_B \cdot \mathbf{c}_B = \mathbf{0} \) (object \( B \) at origin), and the relative offset is \( \mathbf{d}_{AB} = \tilde{\mathbf{c}}_A - \tilde{\mathbf{c}}_B = \tilde{\mathbf{c}}_A \) where \(T_{B}\!\in\!SE(3)\) is the world-to-camera extrinsic transform whose origin is fixed at the centre of object  B and whose axes are aligned with the viewing direction of the image in which B is visible\footnote{More details are discussed in Appendix \ref{appx:camera}}.
% To define ground-truth relations, we compute the position of object $A$ relative to object $B$ from the viewpoint of a frame where $B$ is visible. Let $\mathbf{T}_B$ be the world-to-camera transform for that frame, and let $\mathbf{c}_A, \mathbf{c}_B \in \mathbb{R}^3$ denote the objects' centers in world coordinates. We transform them into the camera coordinate frame:
% \begin{align*}
% \tilde{\mathbf{c}}_A &= \mathbf{T}_B \cdot \mathbf{c}_A \quad \text{(object $A$ in $B$'s frame)} \\
% \tilde{\mathbf{c}}_B &= \mathbf{T}_B \cdot \mathbf{c}_B = \mathbf{0} \quad \text{(object $B$ at origin)} \\
% \mathbf{d}_{AB} &= \tilde{\mathbf{c}}_A - \tilde{\mathbf{c}}_B = \tilde{\mathbf{c}}_A \quad \text{(relative offset)} \\
% \end{align*}

All spatial relations are derived from $\mathbf{d}_{AB}$, which encodes object $A$’s position relative to $B$ in the local frame. This anchors each question to a consistent egocentric perspective.  This is consistent with the directional vector $\mathbf{d}_{AB}$ and the way spatial language is interpreted in egocentric video (e.g., “Is the ottoman to the left of the window?” is framed relative to where the \textit{window} appears as shown in Figure~\ref{fig:teaser}). This design encourages the model to align its internal scene representation with the perspective of the current observation target and camera. To support diverse question types we also extract each object's volume $\mathbf{v}_A, \mathbf{v}_B$  and frame-wise instance maps detailing which objects appear in each frame. 

\paragraph{Question Generation.}
For each object pair $(A, B)$, and their respective reference frames $(f_a, f_b)$, we have now derived all the 3D meta-data to answer spatial questions of different types. We then use predefined templates to come up with QA pairs using this meta data. We then provide this to GPT-4o to paraphrase it to more natural language question answer (QA) pairs. We provide the templates and prompts in Appendix ~\ref{sec:question-templates} and ~\ref{sec:paraphrase-prompt}.

% In addition to the 1.6K scenes used to construct the benchmark, we apply the same pipeline to approximately 10,000 additional scenes to create a large-scale training set. This dataset includes over 3.3 million template-based QA pairs, which can be used for supervised fine-tuning.

\paragraph{Visual Grounding}
To ensure models attend to the correct object instances especially in scenes containing multiple objects of the same type, we provide visual markers similar to Set of Marks \cite{yang2023setofmark}.We project the 3D centers of objects $A$ and $B$ into their respective RGB frames using known camera parameters, and mark these projected centers with visual indicators (e.g., colored hollowed circles). This is demonstrated in Figure ~\ref{fig:teaser}. We find this is an  important factor influencing model performance, with \emph{Marked} baselines outperforming the \emph{Unmarked} counterparts significantly. \S~\ref{sec:results:2d}.
% Without grounding, models may confuse similarly labeled objects (e.g., two chairs), leading to incorrect relational inferences.

\subsection{Dataset Quality Evaluation}
To verify data quality, we sampled 10\% subset of the dataset and perform human evaluation on whether (i) the objects in the question are marked correctly in the video frames (yes/no) and (ii) the answer is accurate in the given context of video frames and the question (yes/no). We found that 99\% of examples were correctly annotated with the appropriate object markers indicating the validity of our meta-data. Further, 96\% of questions were relevant to the provided scene and 94\% of answers were accurate.  \footnote{Appendix ~\ref{sec:crowdsourcing} provides more details}.

\section{Evaluating with 3D Cues}
\label{sec:3dcontext}
\emph{Can a vision-language model reason more effectively when provided with an explicit 3D scene representation, rather than relying on pixel-level inference alone?}
Visual object marking reduces referential ambiguity, but understanding spatial relationships between objects observed in disjoint frames often requires more than isolated 2D snapshots. To investigate this, we evaluate models under two types of 3D augmentations: \textbf{sparse} and \textbf{full 3D context}.
\subsection{Sparse 3D Cues}
\label{subsec:sparse3d}
Sparse 3D cues provide realistic, test-time signals that offer partial information about the scene's spatial layout. We introduce two forms:
\newline\noindent
\textbf{Textual Trajectory Cue.}
We encode the camera trajectory as a sequence of  positions $\mathbf{p}_t = (x_t, y_t z_t)$. This text-based representation reflects how the camera traverses the scene.
\newline\noindent
\textbf{Top-down Cue.}
 We generate a bird’s-eye view (BEV) rendering of the scene. Built from RGB-D and instance segmentation data, the BEV provides a top-down visualization of the scene. It captures the spatial geometry that is not easily inferred from RGB frames alone. For each question, we render a targeted BEV image that highlights the relevant sub-trajectory, from the frame where object A appears to the frame where object B is visible. These visual cues are derived from the full 3D reconstruction but presented in a 2D RGB image that can be readily processed by modern VLMs. For example, in Figure ~\ref{fig:teaser}, the BEV image shows the top down view of the scene, along with a trajectory involving the two objects in the question - ottoman and floor mat. Additional details on generation and prompting with BEV images are in Appendix ~\ref{sec:bev-algo} and ~\ref{sec:bev-prompt}.

\subsection{Full 3D Context}
\label{subsec:full3d}
In addition to evaluating models under sparse cues, we introduce an oracle setting where the model receives dense, ground-truth spatial metadata. This \textbf{full 3D context} includes the precise 3D coordinates of object centers in a global reference frame $C_A = (x_A, y_A, z_A)$ and $C_B = (x_B, y_B, z_B)$, as well as their physical dimensions or bounding box volumes $V_A$ and $V_B$. This representation encodes the metric spatial relationships underlying the correct answer to each question. We use this setting to approximate an \textbf{upper bound on spatial reasoning performance}, isolating reasoning limitations from perceptual errors. While one might consider using predicted 3D detections instead, existing 3D detectors remain brittle, especially in egocentric video—due to occlusions, limited annotations, and poor generalization. Ground-truth cues thus serve as a clean scaffold to evaluate whether failures arise from missing information or from an inability to integrate and reason over spatial geometry.

% Specifically, for each question involving a pair of objects $(A, B)$, we provide, The 3D coordinates of object centers in a global reference frame: $C_A = (x_A, y_A, z_A)$ and $C_B = (x_B, y_B, z_B)$. The physical dimensions or bounding box volumes $V_A$ and $V_B$ of objects $A$ and $B$.

\section{Evaluation Setup}
\vspace{0.5em}
\noindent\textbf{Metric.}
Our dataset follows a similar structure to OpenEQA \cite{majumdar2024openeqa}, where each example consists of an open-ended question grounded in a visual or embodied context, with answers provided in natural language. Due to this alignment in task formulation and answer format (one line open-vocabulary answers), we adopt the LLM-Match metric proposed by OpenEQA to evaluate model predictions. This metric employs a large language model to rate the semantic similarity between predicted and reference answers on a 1–5 scale, providing a more reliable measure for open-ended, free-form QA than conventional string-matching methods. We normalize these scores to the [0, 1] range and report them as percentages. Model and prompt for LLM-Match are detailed in Appendix~\ref{sec:llm-match}.

\vspace{0.5em}
\noindent\textbf{Models.}
We evaluate both closed-source and open-source VLMs. Closed-source models include {GPT-4o} accessed via public API. Open-source models include {LLaVA-Next-Video} (7B), LLava-Video (7B), {InternVL} (8B and 38B) and  {Qwen-VL} (72B). For each model, we standardize the prompt format and provide a sequence of video frames along with the question. We prompt all models to provide a chain-of-thought (CoT) followed by the actual answer. Refer to Appendix ~\ref{sec:traj-prompt} for the prompts.
These models are selected based on their strong performance on recent video reasoning benchmarks such as VideoMME\cite{fu2024video} and their ability to process multi-frame inputs effectively. 

\vspace{0.5em}
\noindent\textbf{Human Performance.}
 We randomly sample a subset of 600 questions for estimating human performance on \datasetname{}. Three human evaluators independently answer each question, and their performance is evaluated using the same LLM-Match metric. Appendix ~\ref{sec:crowdsourcing} provides further details on crowdsourcing.

\section{Empirical Evaluation}
\label{sec:results}
We structure our evaluation around three core research questions designed to assess how well VLMs reason about spatial relationships in egocentric video, particularly when the objects in question never co-occur in the same frame.
\begin{enumerate}[label=\large\protect\textcircled{\small\arabic*}]
    \item \textbf{RQ1:} \textbf{2D-Only --} Can VLMs reason about spatial relations using only 2D egocentric video, and how does visually disambiguating object references affect performance?

    \item \textbf{RQ2:} \textbf{Effect of 3D Cues --} Does providing explicit 3D spatial cues, either linguistically or visually, enhance model performance?

    \item\textbf{RQ3:} \textbf{Failure Modes --} What factors (e.g., object distance) make spatial reasoning particularly challenging for current models?
\end{enumerate}

\subsection{\textbf{RQ1: 2D-Only — Does visual disambiguation improve spatial reasoning?}}
\label{sec:results:2d}
% \begin{table}[ht]
% \centering
% \small
% \rowcolors{4}{gray!4}{white}
% \renewcommand{\arraystretch}{1.2}
% \resizebox{0.9\columnwidth}{!}{
% \begin{tabular}{lcc}
% \toprule
% \textbf{Model} & \textbf{Objects Unmarked} & \textbf{Objects Marked} \\
% \midrule
% \multicolumn{3}{l}{\textit{Closed-Source Models}} \\
% GPT-4o & 62.88 & \textbf{65.60} \\
% % Gemini 1.5 Pro & -- & tbd \\
% \midrule
% \multicolumn{3}{l}{\textit{Open-Source Models}} \\
% LLaVa-NeXT-Video 7B & 48.44 & 49.92 \\
% LLaVA-Video 7B      & 56.40 & 63.30 \\
% InternVL3-8B        & 53.16 & 61.80 \\
% InternVL3-38B       & 54.60 & 62.30 \\
% Qwen 2.5-72B        & 55.06 & \textbf{64.31} \\
% \midrule
% Human Performance   & tbd   & -- \\
% \bottomrule
% \end{tabular}}
% \caption{Normalized LLM-Match (\%) across models in two evaluation settings: \textbf{Unmarked} (no object highlighting) and \textbf{Marked} (referenced objects visually indicated). Closed- and open-source models are grouped separately. Human performance is reported for reference.
% }
% \label{tab:llm-match-overall}
% \end{table}

\begin{figure*}[t]
  \centering

  %--- Subfigure A: Table ----------------------------
  \begin{subfigure}[t]{0.40\textwidth}
    \centering
    \vspace{0pt}  % fix vertical alignment
    \parbox[t]{\linewidth}{
      \rowcolors{4}{gray!4}{white}
      \renewcommand{\arraystretch}{1.2}
      \resizebox{\linewidth}{!}{
\begin{tabular}{lccc}
  \toprule
  \textbf{Model} & \textbf{Unmarked} & \textbf{Marked} & \textbf{$\Delta$} \\
  \midrule
  \multicolumn{4}{l}{\textit{Closed-Source Models}} \\
  GPT-4o & 62.88 & \textbf{65.60} & +2.72 \\
  \midrule
  \multicolumn{4}{l}{\textit{Open-Source Models}} \\
  LLaVa-NeXT-Video 7B & 48.44 & 49.92 & +1.48 \\
  LLaVA-Video 7B      & 56.40 & 63.30 & +6.90 \\
  InternVL3-8B        & 53.16 & 61.80 & +8.64 \\
  InternVL3-38B       & 54.60 & 62.30 & +7.70 \\
  Qwen 2.5-72B        & 55.06 & \textbf{64.31} & \textbf{+9.25} \\
  \midrule
  Human Performance   & --   & 93.96    & -- \\
  \bottomrule
\end{tabular}
      }
      \subcaption{Normalized LLM-Match (\%) across models in two evaluation settings: \textbf{Unmarked} (no object highlighting) and \textbf{Marked} (referenced objects visually indicated).}
      \label{tab:llm-match-overall}
    }
  \end{subfigure}
  \hfill
  %--- Subfigure B: Figure ----------------------------
  \begin{subfigure}[t]{0.56\textwidth}
    \centering
    \vspace{0pt}
    \includegraphics[width=\linewidth]{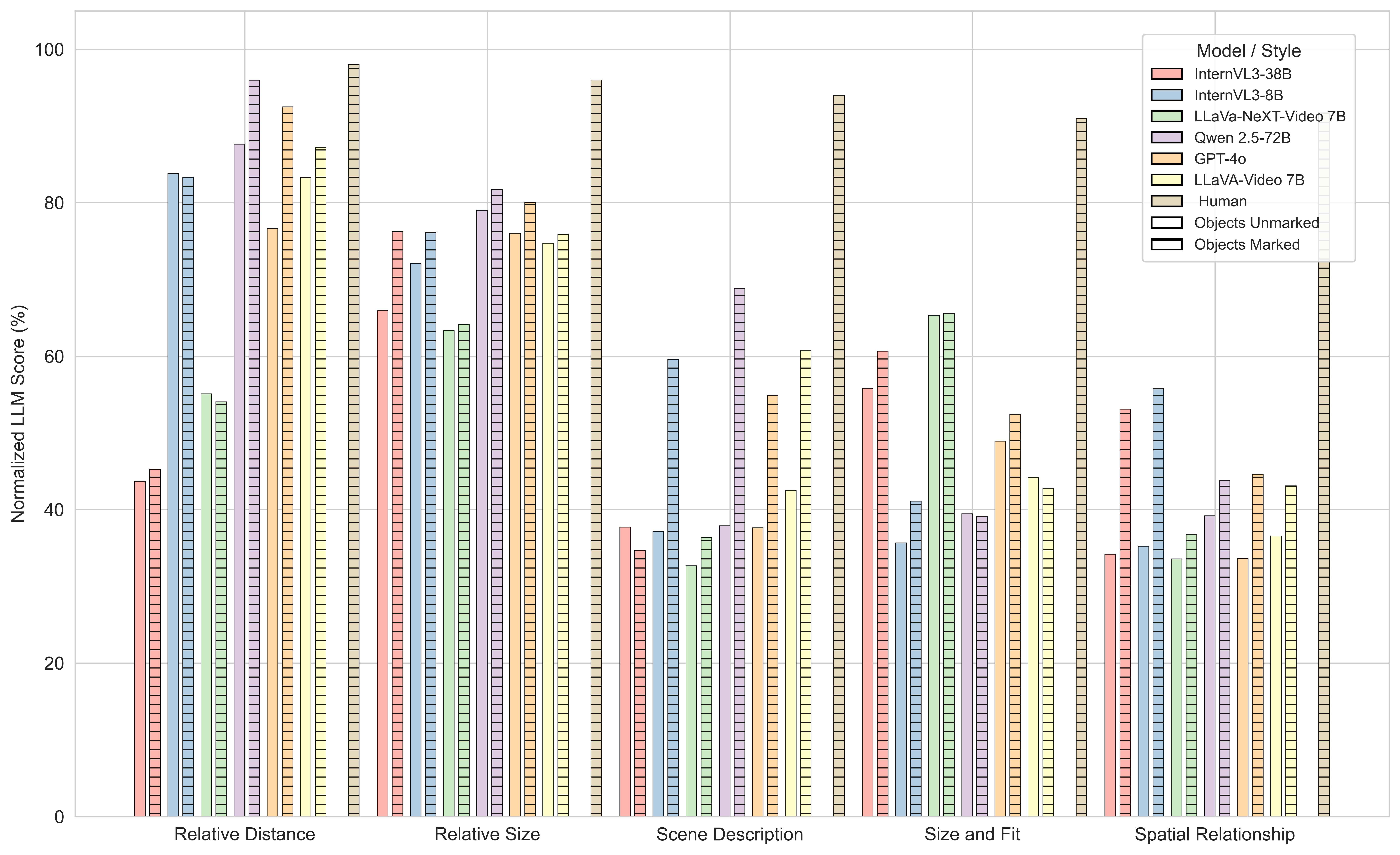}
    \subcaption{Normalized LLM-Match (\%) across spatial categories}
    \label{fig:rq1}
  \end{subfigure}

  \caption{{RQ1: 2D-Only} — Does visual disambiguation improve spatial reasoning: Comparison of LLM performance with and without visual object disambiguation. Left: overall accuracy across models. Right: performance by spatial category.}
\end{figure*}
In the 2D-Only setting, we provide the VLMs with video frames and the question and investigate their performance out-of-the box \textit{Unmarked} vs \textit{Marked
}. The \textit{Marked} setting refers to the visual grounding setup described in ~\ref{sec:dataset}, where objects relevant to the question are marked with red hollow circle to visually guide the models.

Figure~\ref{tab:llm-match-overall} summarizes overall score and Figure~\ref{fig:rq1} presents a breakdown of model performance across spatial categories. Closed-source models like \textbf{GPT-4o} exhibit notable spatial reasoning abilities out-of-the-box, reaching an accuracy of 62.88\% even without visual marking of objects. With objects disambiguated, we see an improvement of nearly 3\% over the \textit{Unmarked} setting.  Open-source models display greater sensitivity to visual prompting with markers, with improvements of approximately \textbf{7--9\%} between the \textit{Unmarked} and \textit{Marked} settings. For instance, LLaVA-Video 7B, Qwen 2.5-72B, and InternVL3-8B all show substantial gains in response to object highlighting. These jumps suggest that even simple referential cues provide a strong inductive signal, helping models resolve ambiguous object references and reason more effectively in egocentric scenes.

 As shown in Table~\ref{tab:llm-match-overall}, humans achieve a normalized LLM-Match score of \textbf{93.96\%}, outperforming all models by a wide margin. Notably, even the best-performing model—GPT-4o with visual markers—lags behind by over 28 percentage points. This performance gap persists across all spatial categories (Figure~\ref{fig:rq1}
Beyond overall scores, the per-category analysis in Figure~\ref{fig:teaser} reveals key trends. Most models perform well on categories such as \textbf{Relative Distance} and \textbf{Relative Size}, where spatial relationships are often visually salient and co-visible within frames and common notions understood in the language domain, such as a vase is likely smaller than a couch. However, all models, regardless of architecture or training data, struggle in more complex categories like \textbf{Size and Fit} and especially \textbf{Spatial Relationship}, which require multiframe integration.

\vspace{0.5em}
\noindent
\textit{Takeaway: Explicit visual disambiguation significantly boosts model accuracy, particularly for open-source models. However, a substantial gap remains between the best-performing models and humans, highlighting limitations in current spatial reasoning and multi-frame integration.}

\subsection{\textbf{RQ2: 3D Cues — Do 3D Cues Improve Reasoning in VLMs}}
\label{sec:results:3d}
% \begin{table}[t!]
% \footnotesize
% \centering
% \rowcolors{3}{gray!4}{white}  % subtle zebra striping
% \renewcommand{\arraystretch}{1.2}  % more vertical spacing
% \resizebox{\columnwidth}{!}{
% \begin{tabular}{@{}llcc@{}}
% \toprule
% \textbf{Category} & \textbf{Input Type} & \textbf{Qwen 2.5-72B} & \textbf{GPT-4o} \\
% \midrule
% \multicolumn{4}{l}{\textit{\textbf{(1) Baselines}}} \\
% -- & No Marking & 55.06 & 62.88 \\
% -- & Visual Marking Only & 64.31 & 65.60 \\
% \midrule
% \multicolumn{4}{l}{\textit{\textbf{(2) Realistic 3D Context (Test-time Cues)}}} \\
% Linguistic & 3D Trajectory (Text) & xx& 67.60\\
% % Visual & Full BEV (Whole Scene) & xx & xx \\
% Visual & BEV (A$\rightarrow$B Sub-Trajectory Only) & 66.64  & 68.23 \\
% \midrule
% \multicolumn{4}{l}{\textit{\textbf{(3) Upper Bound (Ground Truth Info)}}} \\
% Linguistic & \makecell[l]{Direct Spatial Metadata \\ (centers, volumes, trajectories)} & - & 83.2 \\
% \bottomrule
% \end{tabular}
% }
% \caption{
% \small
% Effect of visual and linguistic 3D context on model performance. Baselines reflect zero-shot settings. Realistic cues are test-time injectables; the upper bound simulates oracle access to ground-truth geometry. Normalized LLM Match (\%) is reported.
% }
% \label{tab:3d_context_results}
% \end{table}

\begin{figure*}

  \begin{subfigure}[t]{0.48\textwidth}
    \centering
    \vspace{0pt}  % fix vertical alignment
    \parbox[t]{\linewidth}{
      \rowcolors{3}{gray!4}{white} 
      \renewcommand{\arraystretch}{1.2}
      \resizebox{\linewidth}{!}{
\begin{tabular}{@{}llcc@{}}
\toprule
\textbf{Modality} & \textbf{Input Type} & \textbf{Qwen 2.5-72B} & \textbf{GPT-4o} \\
\midrule
\multicolumn{4}{l}{\textit{\textbf{(1) Baselines}}} \\
-- & No Marking & 55.06 & 62.88 \\
-- & Visual Marking Only & 64.31 & 65.60 \\
\midrule
\multicolumn{4}{l}{\textit{\textbf{(2) Sparse 3D Context (Test-time Cues)}}} \\
Text & 3D Trajectory (Text) & 66.21 & 67.60\\
% Visual & Full BEV (Whole Scene) & xx & xx \\
Image & BEV (A$\rightarrow$B Sub-Trajectory Only) & 66.64  & 68.23 \\
\midrule
\multicolumn{4}{l}{\textit{\textbf{(3) Full 3D Context (Ground Truth Upperbound)}}} \\
Text & \makecell[l]{Direct Spatial Metadata \\ (centers, volumes, trajectories)} & - & 83.2 \\
\bottomrule
\end{tabular}      }
      \subcaption{Effect of sparse and full 3D context on model performance. Normalized LLM Match (\%) is reported.}
      \label{tab:3d_context_results}
    }
  \end{subfigure}
  \hfill
  %--- Subfigure B: Figure ----------------------------
  \begin{subfigure}[t]{0.48\textwidth}
    \centering
    \vspace{0pt}
    \includegraphics[width=\linewidth]{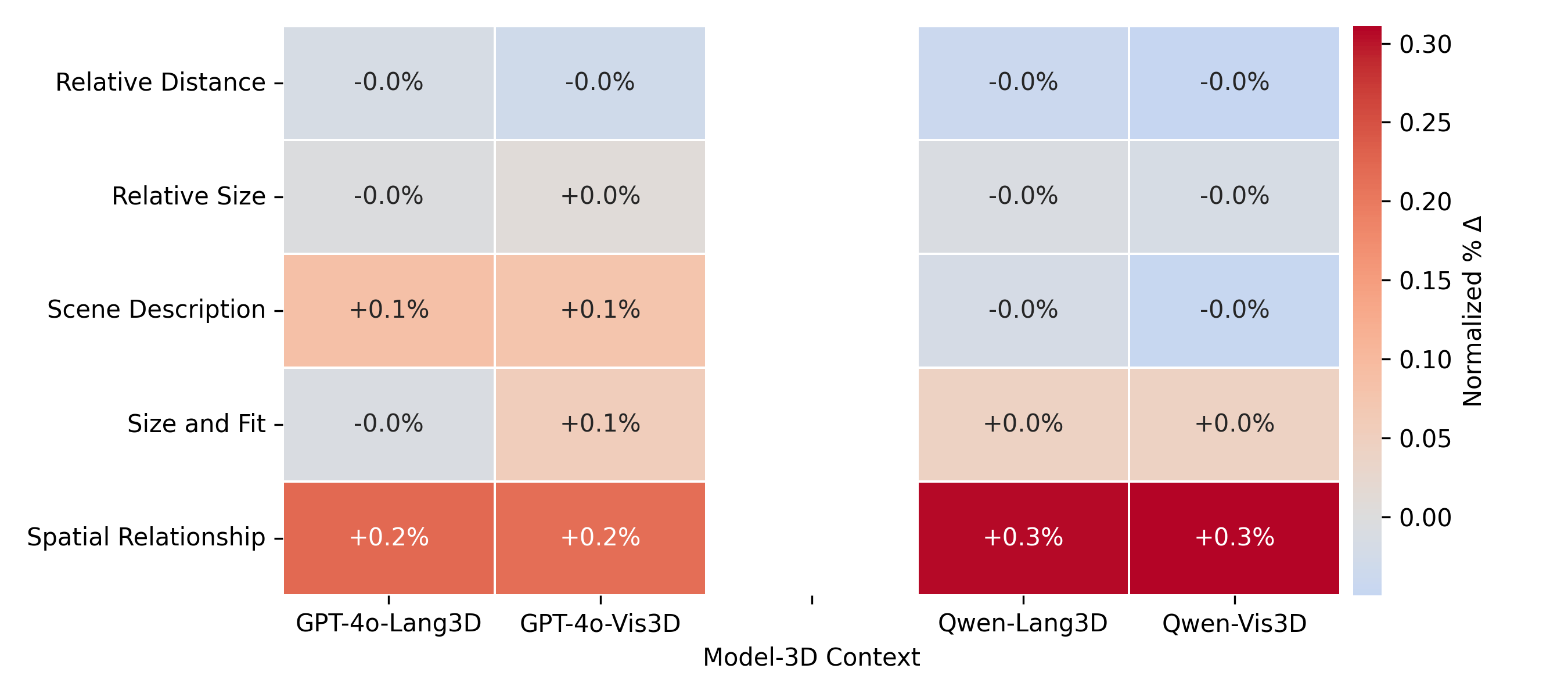}
    \subcaption{$\Delta$ from base to 3D-augmented prompts for GPT-4o and Qwen - Gains are seen in Scene Description and Spatial Relations.}
    \label{fig:rq2}
  \end{subfigure}

  \caption{{RQ2: 3D Cues} — Do 3D cues improve performance: Comparison of LLM performance with visual and linguistic 3D cues. Left: Overall performance with different cues. Right: Performance gained by spatial category.}
\end{figure*}
 Table~\ref{tab:3d_context_results} quantifies the effect of injecting 3D spatial information into VLMs. As described in \S~\ref{sec:3dcontext}, we assess two settings: a \textit{sparse augmentation} setting where test-time 3D cues (e.g., A$\rightarrow$B bird's-eye view with sub-trajectories or language-based trajectories) are provided, and a \textit{Full 3D context} setting with access to ground truth metadata (e.g., object volumes and positions). We compare it against a \textit{baseline} with object markings only.

Across both GPT-4o and Qwen-72B, introducing sparse 3D cues consistently improves performance: BEV-based visual augmentation boosts GPT-4o by 2.6\%, while linguistic descriptions yield slightly smaller gains. For Qwen, both visual and linguistic 3D inputs lead to $\sim$4\% improvement.

Providing direct access to spatial metadata, such as object centers, volumes, and trajectories leads to a dramatic 18\%  jump in performance for GPT-4o. This upper bound reflects the advantage of structured geometry, where reasoning reduces to direct comparisons of object centers or volumes.  Our results suggest that the bottleneck lies not in reasoning over 3D spatial inputs, but in constructing accurate 3D representations from sparse or 2D observations. 

% This highlights both the limitations of current perceptual inputs and the potential of structured scene representations for enhancing spatial understanding.

\noindent\textbf{Category-wise Gains.} To further dissect these improvements, Figure~\ref{fig:rq2} (right) shows the normalized performance change ($\Delta$) in five categories of spatial reasoning. The gains are most pronounced in \textit{Scene Description} and \textit{Spatial Relationship} questions, which demand a global understanding of object layout and egocentric traversal.

\vspace{0.5em}
\noindent
\textit{Takeaway: Realistic 3D cues offer modest gains especially for relational reasoning. The strong boost from ground-truth metadata highlights that the bottleneck lies in constructing accurate 3D representations from sparse cues or visual signals. 
}

\subsection{\textbf{RQ3: Failure Modes}}
\label{sec:results:failure}
\paragraph{Distance Based Failure Modes.}
\begin{figure}
  \includegraphics[width=\columnwidth]{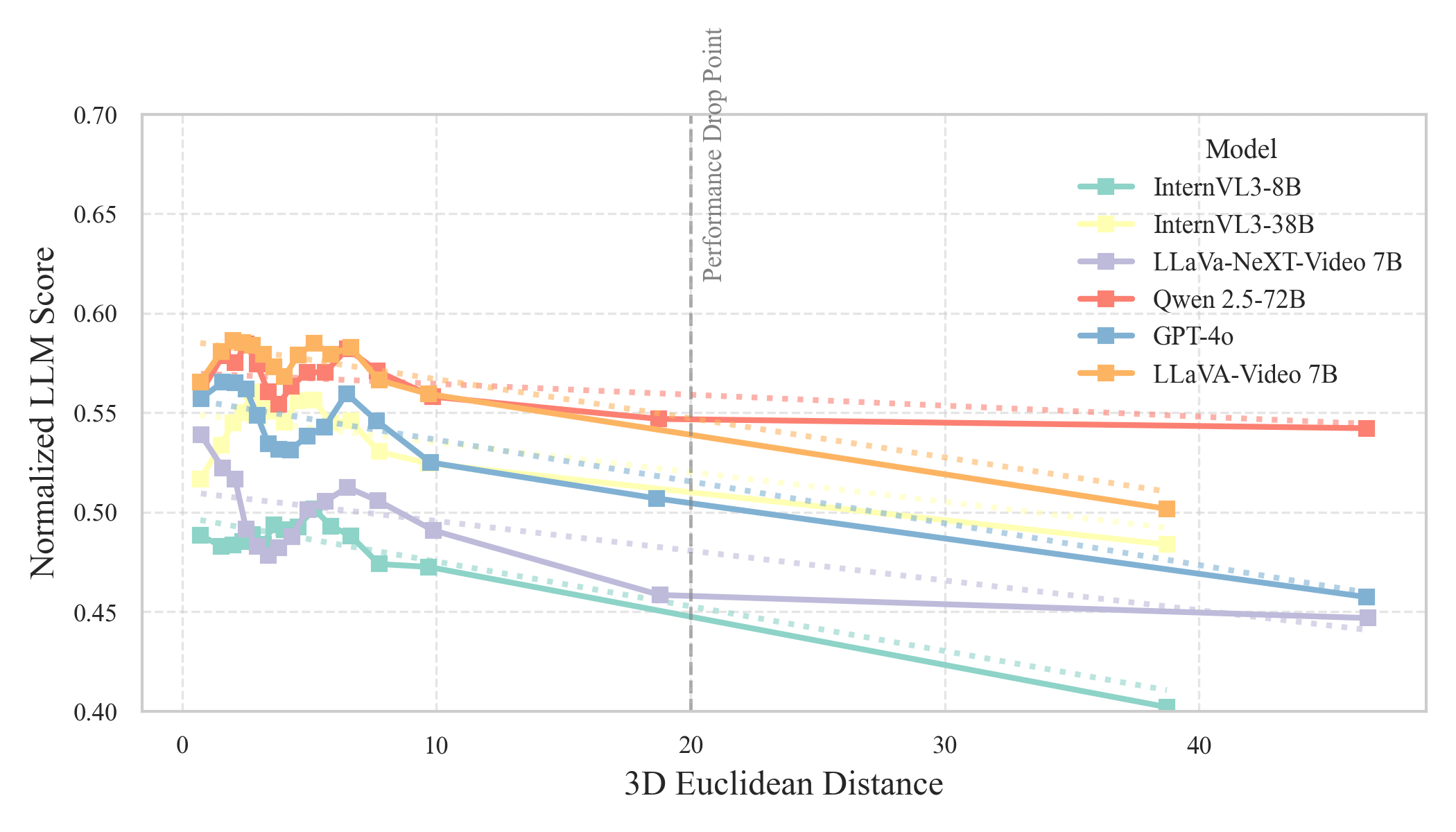}
  \caption{RQ3: Failure Modes: Model performance declines with increasing Euclidean distance between objects.}
  \label{fig:distance_degradation}
\end{figure}
Figure~\ref{fig:distance_degradation} illustrates how model performance varies with the 3D Euclidean distance between object pairs. Across all models, accuracy declines as spatial separation increases. This trend is especially pronounced for models like InternVL3-8B, which show sharper degradation. Larger models such as Qwen 2.5-72B and GPT-4o seem slightly more robust to moderate distances. However, performance drops significantly once the separation exceeds ~20 metres, typically corresponding to objects located in different rooms.
\paragraph{Success vs. Failure of explicit cues}
\begin{figure*}
  \includegraphics[width=\textwidth]{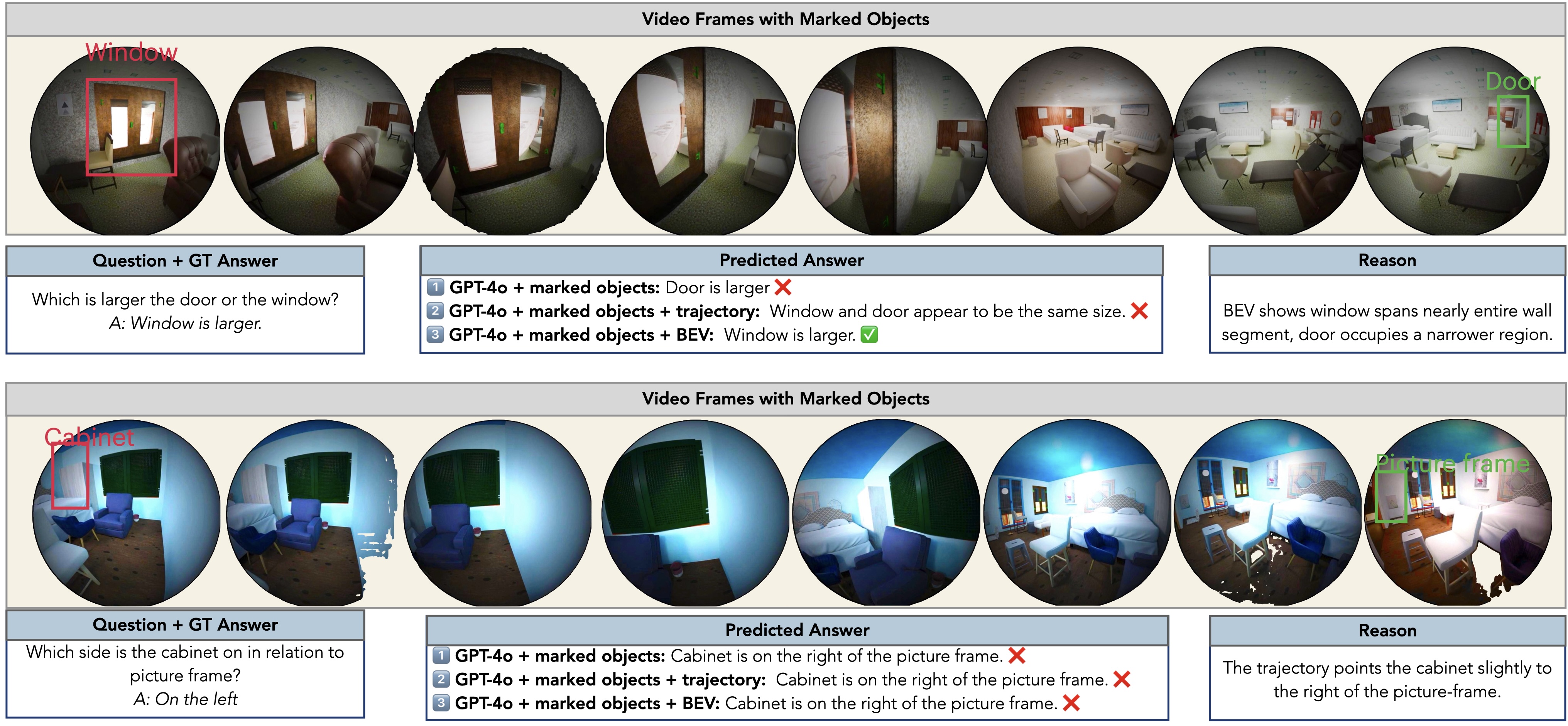}
\caption{RQ3: Failure Modes. GPT-4o model predictions under different visual contexts: (i) marked objects, (ii) + trajectory, (iii) + BEV. Top: only BEV resolves spatial fit. Bottom: all models fail.}

  \label{fig:qual}
\end{figure*}
Figure~\ref{fig:qual} compares the predictions of GPT-4o-based models under three visual grounding settings: (i) marked objects only, (ii) marked objects with egocentric trajectory, and (iii) marked objects with bird’s-eye view (BEV) context.
In the first row, the model must compare the size of the door and window across disjoint views. Without spatial grounding, GPT-4o misjudges the door as larger or expresses uncertainty. With BEV input, it correctly identifies the window as larger, matching ground truth,  indicating that access to top-down scene views helps in understanding relative sizes.
In the second example, the question concerns the directional spatial relation of whether the cabinet is to the right of the picture frame. Here, all three model variants answer wrongly, despite providing 3D context such as trajectories and the BEV.
\paragraph{Error analysis of Failure Modes of explicit cues.}
Across the full test split, adding BEV inputs yields a 2.5\% improvement over the baseline for GPT-4o. We manually examined 50 examples where BEV maps led to improved performance. In 81\% of these, the model’s chain-of-thought (CoT) explicitly referenced the new cue e.g., ``the BEV shows the window spans a wider area”. However, an analysis of 50 failure cases with BEV inputs reveals several distinct error modes. In 18\% of failures, the model mentioned the BEV but misinterpreted spatial relations, such as confusing directions or swapping the start and end of a trajectory. Another 17\% ignored the BEV entirely without referencing the added spatial context. In 24\% of cases, the model produced hallucinated geometry, making confident yet unfounded claims about object size, position, or visibility.
We also observed that 15\% of failures involved misalignment between egocentric views and BEV maps, where the model failed to correctly associate objects across views.  These findings suggest that while BEV maps provide useful spatial context, their effectiveness depends on the model’s ability to align, interpret, and selectively attend to the added modality. 

% Gains occur when BEV cues are meaningfully integrated into the reasoning process; failures, by contrast, reveal persistent gaps in multimodal grounding and spatial abstraction.
% Across the full test split, adding BEV or trajectory inputs yields a \(\approx\)2.5 \% boost over the \textsc{Marked} baseline for GPT4o.  To see whether this gain manifests itself in the reasoning of the model, we manually examined 50 of the \textit{improved} examples.  In 81\% of them, the CoT explicitly referenced the new cue; e.g., as shown in Figure~\ref{fig:qual}, \emph{the BEV shows the window spans a wider area}”.  For the much larger set of \textit{failure} cases with 3D cues, CoTs paint a different picture: (1) 44 \% do not leverage the cues to perform reasoning, falling back on frame-based heuristics;  (2) 33 \% mention the cue but misinterpret it (e.g., swapping start/end of the sub-trajectory); and  (3) 23 \% produce hallucinated geometry despite the extra input. These patterns indicate that sparse 3D context \emph{helps} when the model can correctly align it with the egocentric frames, but this task is challenging and may be introducing new sources of error.
\section{Related Work}
\begin{table*}[t]
\centering
\small
\resizebox{\textwidth}{!}{
\begin{tabular}{lcccccc}
\toprule
\textbf{Dataset} & \textbf{Viewpoint} & \textbf{Reasoning Type} & \textbf{Object Co-visibility} & \textbf{Temporal Scope} & \textbf{3D Context} & \textbf{Task Format} \\
\midrule
\textbf{SpatialRGPT-Bench}~\cite{cheng2024spatialrgpt} & Mixed (Indoor/Outdoor/Simulated) & Spatial (2D/3D) & Varies & Single-frame & Yes & Gen \\
\textbf{EmbSpatial-Bench}~\cite{Newcombe2024Spatial} & Egocentric & Spatial (6 relations) & Mostly Yes & Multi-frame & No & MCQ \\
\textbf{OpenEQA}~\cite{majumdar2024openeqa} & Egocentric & Spatial / Commonsense & Mostly Yes & Multi-frame & No & MCQ \\
\textbf{VSI-Bench}~\cite{yang2024think} & Egocentric & Spatial (3D) & Yes & Multi-frame & No & MCQ \\
\midrule
\textbf{\datasetname{} (Ours)} & Egocentric & Spatial (3D) & \textbf{No} & \textbf{Multi-frame} & \textbf{Optional} & Gen \\
\bottomrule
\end{tabular}
}
\caption{Comparison of recent video benchmarks closely related to \datasetname{}. Our benchmark, uniquely emphasizes spatial reasoning without object co-visibility and evaluates models by incorporating 3D scene structure.}
\label{tab:dataset_comparison}
\end{table*}
\subsection{Egocentric 3D Understanding}
Reasoning about space from an egocentric viewpoint is fundamental to embodied intelligence, enabling agents to navigate, manipulate, and interact with their environment~\cite{Ruggiero2009SpatialMemory}.  Recent work has advanced 3D spatial understanding from egocentric inputs across several fronts. EgoGaussian~\cite{Zhang2024EgoGaussian} reconstructs static scenes using 3D Gaussian splatting from monocular egocentric videos. EgoSplat~\cite{Park2025EgoSplat} extends this approach with open-vocabulary capabilities and EgoSG~\cite{Zhang2024EgoSG} proposes building 3D scene graphs from egocentric footage. Large-scale egocentric datasets such as Ego4D~\cite{Grauman2022Ego4D} and EPIC-KITCHENS~\cite{Damen2018EPICKITCHENS, Damen2020EPICKITCHENSTPAMI} have catalyzed progress in this area, supporting tasks like action recognition, spatial localization, and object interaction.
\subsection{Video Reasoning Benchmarks for VLMs}
The growing capabilities of multimodal large language models (MLLMs) ~\cite{hurst2024gpto,team2024gemini,li2024llavaov,wang2024qwen2vl,zhang2024llavanextvideo,xue2024longvila} have motivated the development of benchmarks to evaluate video understanding. Several recent efforts target third-person or general-purpose video reasoning, such as MVBench~\cite{li2024mvbench}, VideoBench~\cite{ning2023videobench}, TempCompass~\cite{liu-etal-2024-tempcompass}, and Video-MME~\cite{fu2024video}, which evaluate temporal ordering, event reasoning, and modality alignment. MotionBench~\cite{hong2025motionbenchbenchmarkingimprovingfinegrained} focuses on fine-grained motion understanding.

In contrast, our work focuses on egocentric and embodied settings, where spatial reasoning is grounded in an agent-centric reference frame. Benchmarks such as EgoSchema~\cite{mangalam2023egoschema} and EgoSpeak~\cite{kim-etal-2025-egospeak} evaluate language grounding and QA in egocentric contexts. OpenEQA~\cite{majumdar2024openeqa} and VSI-Bench~\cite{yang2024think} are most relevant to our focus on spatial reasoning. OpenEQA has spatial questions that often involve co-visible objects. VSI-Bench tasks involve generating cognitive maps from egocentric RGB inputs, but do not enforce object non-co-visibility and omit explicit 3D priors, effectively asking models to infer spatial structure without access to geometry. We explicitly enforce disjoint-frame spatial reasoning and optionally provide 3D structure, enabling a more controlled evaluation of spatial reasoning capabilities in VLMs. We show a comparison against spatial reasoning benchmarks in Table ~\ref{tab:dataset_comparison}.

\subsection{Spatio-Temporal Reasoning with MLLMs}
Recent work has highlighted the limitations of MLLMs~\cite{hurst2024gpto, team2024gemini, li2024llavaov, wang2024qwen2vl, zhang2024llavanextvideo, xue2024longvila} in fine-grained spatial and temporal reasoning and worked on methods to improve them. \newcite{Zhang2025CallNewRecipes} argue that architectural scaling alone is insufficient, advocating for spatially-aware objectives, structured supervision, and better positional encodings.  More recent work has explored concrete strategies to improve VLM spatial reasoning. \newcite{Liao2025ImprovedVisualSpatial} propose Group Relative Policy Optimization (GRPO), which fine-tunes VLMs using spatially grounded supervision and demonstrates substantial gains on spatial benchmarks. Parallel research explores methods such as coarse correspondence supervision~\cite{liu2024coarsecorrespondenceelicit3d}, and unified objectives for spatial understanding~\cite{yang2024virl, chen2024spatialvlm, cheng2024spatialrgpt, cai2024spatialbot, zhu2024llava3d}. 
\section{Conclusion}
We introduce \datasetname{}, a new benchmark designed to evaluate spatial reasoning in egocentric video where objects appear accors multiple frames. Through controlled experiments varying the availability of visual, textual, and 3D cues, we reveal key limitations of current vision-language models (VLMs). Despite improvements from simple object marking and sparse 3D augmentations, models still struggle with tasks that require integrating information across disjoint frames.  We hope our benchmark inspires the development of models that can internalize 3D priors, map 3D scenes, and robustly reason over complex egocentric environments.
\section*{Limitations}
\paragraph{Sythetic Data.} \datasetname{} is built using the Aria Synthetic Environments (ASE) dataset. While ASE is designed to be realistic and offers controllability and ground truth, the direct transferability of findings to the complexities and noise of real-world, unconstrained egocentric videos remains an open question. 

\paragraph{Question Types.}\datasetname{} spans a range of fundamental spatial relations. However, spatial reasoning includes other complex aspects or nuanced question types such as navigation (e.g, ``How do I find my way back from the living room to the kitchen?"), layout (e.g, ``How is the furniture organized in the room?") and compositional spatial reasoning (e.g.,``How do I clean my room?")

\paragraph{Level of ``Sparsity'' and Realism.} The sparse cues used, while not providing the full 3D context -- especially the BEV built from RGB-D and instance segmentation -- are relatively processed and structured.  In many real-world embodied AI scenarios, sparse 3D information might be noisier. It may not be representative of the more incomplete, or ambiguous nature of sparse 3D information often encountered in real-world scenarios.

\paragraph{3D Cues.} The BEV map, while a useful abstraction, is a 2D projection of the 3D world. It simplifies the vertical dimension and can still have limitations in representing complex multi-level scenes or detailed object shapes from a top-down perspective.

\textit{Future work could address these limitations by incorporating real-world egocentric datasets, expanding the range of spatial question types, and exploring less structured or learned representations of 3D context that more closely mirror the noise and ambiguity found in practical scenarios.}

\clearpage
\bibliography{custom}
\clearpage
\appendix
\appendix
\newpage
\section{Appendix}
\label{sec:appendix}
%%%%%%%%%%%%%%%%%%%%%%%%%%%%%%%%%%%%%%%%%%%%%%%%%%%%%%%%%%%%%%%%%%%%%%%%%%%%%%%%%%%%%%%%%%%%%%%%%%%%%%%%%%%%%%%%%%%%%%%%%%%%%%
% \subsection{Additional Qualitative Examples}
% \label{sec:Qualitative}

\subsection{Detailed Statistics of \datasetname{}}
\label{sec:dataset-stats}
\begin{figure}[ht]
    \centering
    \includegraphics[width=\columnwidth]{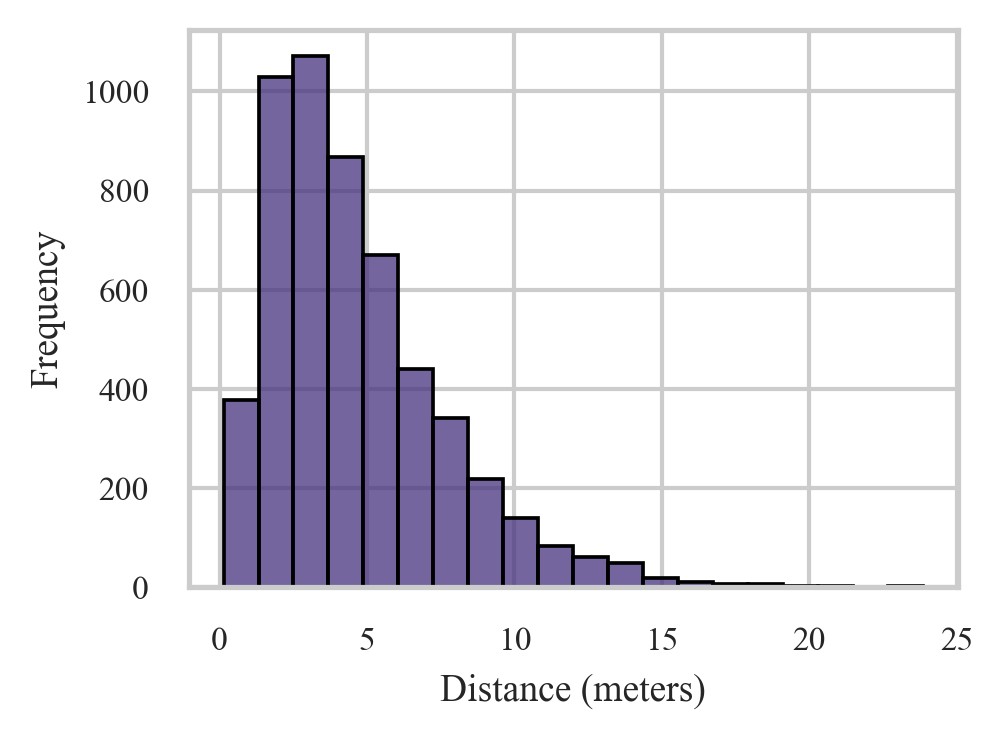}
    \caption{Distribution of spatial distances between object pairs in \datasetname{}. The majority of questions involve objects that are 2–6 meters apart, with a long tail extending up to 20 meters. This highlights the need for long-range spatial reasoning across frames.}
    \label{fig:distance}
\end{figure}

\begin{figure}[ht]
    \centering
    \includegraphics[width=\columnwidth]{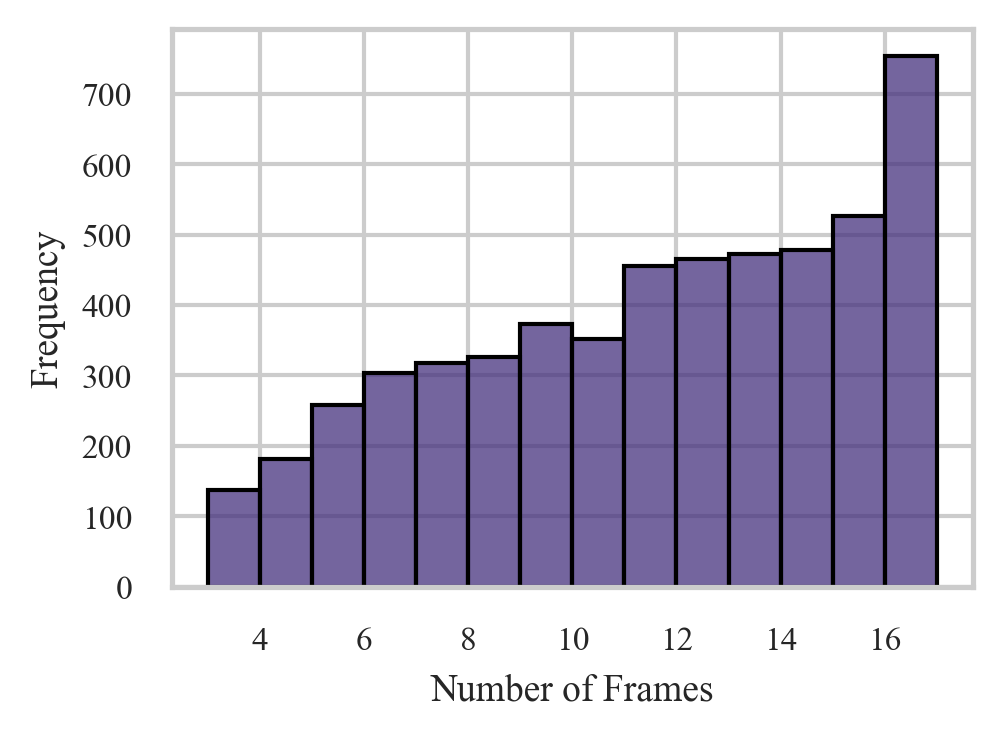}
   \caption{Distribution of the number of frames required to answer each question. Most questions span more than 10 frames, underscoring the need for multi-frame integration and memory over long temporal contexts.}
    \label{fig:frames}
\end{figure}

\begin{figure}[ht]
    \centering
    \includegraphics[width=0.7\columnwidth]{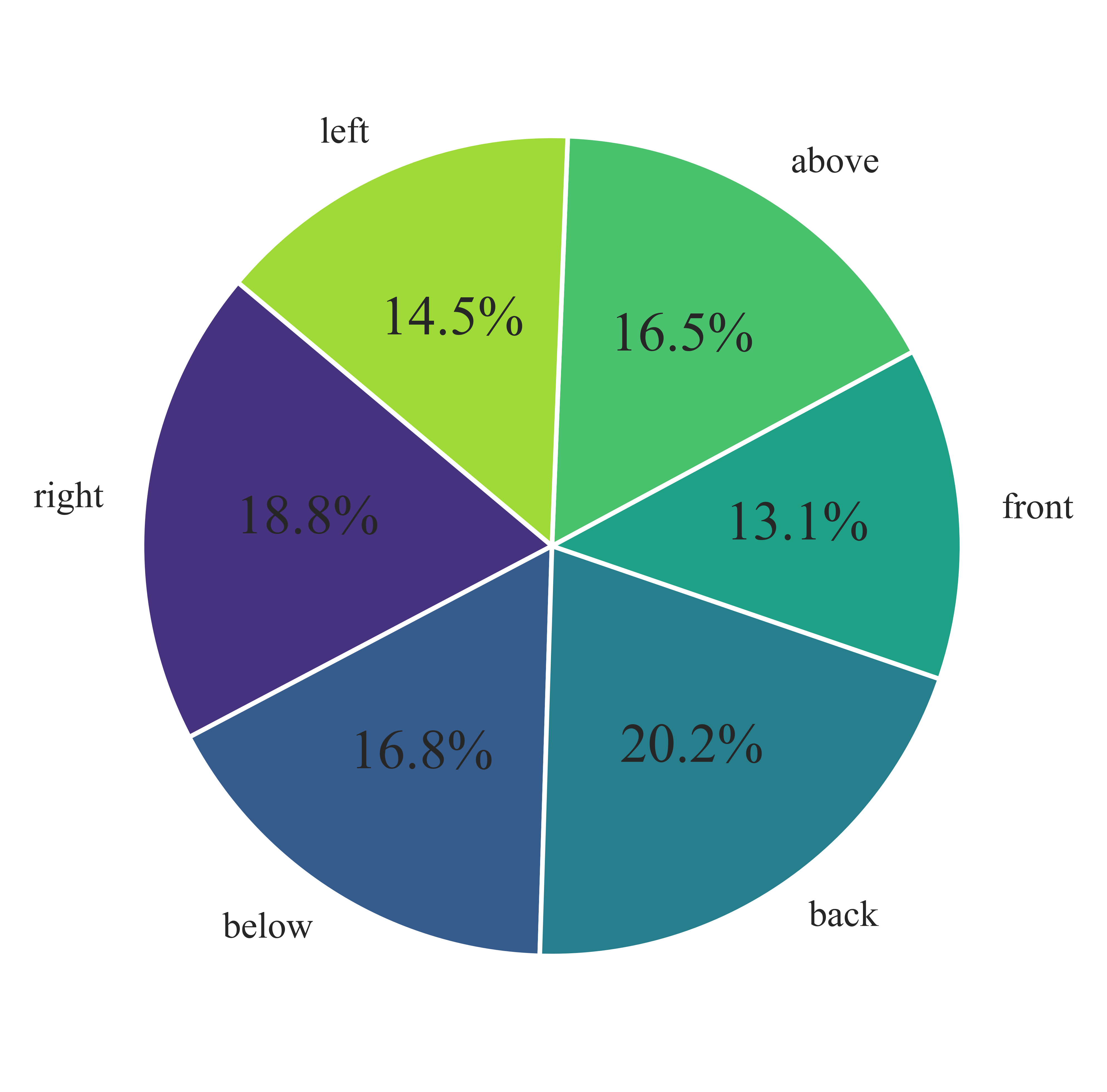}
   \caption{Distribution of directions for questions on spatial relationships.}
    \label{fig:labels}
\end{figure}

Table~\ref{tab:dataset_statistics} summarizes key statistics of \datasetname{}. In figures ~\ref{fig:distance}, ~\ref{fig:frames} we show the distributions of distances between object pairs in terms of 3D distance, number of frames.  In Figure ~\ref{fig:labels} we show the distribution of labels in object relationship questions.

\begin{table}[h]
\centering
\footnotesize
\resizebox{0.7\columnwidth}{!}{
\begin{tabular}{lccc}
\toprule
Scenes & 1668 \\
Total QA pairs & 5399 \\
Unique object pairs (disjoint) & 856 \\
Avg. frames per question & 12 \\
Avg. questions per scene & 4 \\
\bottomrule
\end{tabular}}
\caption{Summary statistics for \datasetname{}.}
\label{tab:dataset_statistics}
\end{table}

\subsection{Templates for \datasetname{}}
\label{sec:question-templates}

To construct natural language questions, we design a set of templates aligned with core spatial reasoning capabilities. These templates reflect the underlying structure of each question type in the dataset and are instantiated with specific object labels (e.g., $ \texttt{object\_a} $, $ \texttt{object\_b} $) based on scene annotations. The resulting questions test models on relational reasoning, physical affordances, and basic object semantics in egocentric video. Some example templates are shown in ~\ref{tab:template_examples}.

\begin{table*}[ht]
\centering
\small
\setlength{\tabcolsep}{10pt}
\renewcommand{\arraystretch}{1.5}
\begin{tabular}{@{}p{3.2cm}p{13.5cm}@{}}
\toprule
\textbf{Category} & \textbf{Question Templates} \\
\midrule
\textbf{Spatial Relationship} &
$\textit{Relative to the } \texttt{object\_a}, \textit{ is the } \texttt{object\_b} \textit{ on the left or right?}$ \\
& $\textit{Is the } \texttt{object\_a} \textit{ vertically above or below the } \texttt{object\_b}\textit{?}$ \\
\midrule
\textbf{Relative Distance} &
$\textit{Are the } \texttt{object\_a} \textit{ and the } \texttt{object\_b} \textit{ within arm's reach of each other?}$ \\
& $\textit{Can you touch both the } \texttt{object\_a} \textit{ and } \texttt{object\_b} \textit{ from one spot?}$ \\
\midrule
\textbf{Size and Fit} &
$\textit{Which object is larger, the } \texttt{object\_a} \textit{ or the } \texttt{object\_b}\textit{?}$ \\
& $\textit{Can the } \texttt{object\_a} \textit{ fit on top of the } \texttt{object\_b}\textit{?}$ \\
& $\textit{Can the } \texttt{object\_a} \textit{ be stacked on the } \texttt{object\_b} \textit{ without falling?}$ \\
\midrule
\textbf{Scene Description} &
$\textit{List the unique objects in the scene?}$ \\
& $\textit{What is the function of the } \texttt{object\_a} \textit{ in this scene?}$ \\
\bottomrule
\end{tabular}
\caption{Representative question templates across reasoning categories in \datasetname{}. Each template is grounded in egocentric visual context and instantiated with object pairs sampled from real scenes.}
\label{tab:template_examples}
\end{table*}

\subsection{Prompt for GPT-4o for Question Paraphrasing}
\label{sec:paraphrase-prompt}
To diversify question phrasing while preserving meaning, we use GPT-4o to generate paraphrased variants of our base templates. The following system prompt is used:

\begin{quote}
\ttfamily
Given a spatial question in natural language, your task is to rephrase it in a different and natural manner while preserving its meaning. The rephrased question should not alter the answer to the question. Do not change the objects mentioned. Avoid yes/no inversion.\\

\textbf{Input:} [Question] [Answer]\\
\textbf{Output:} [Paraphrased Question]
\end{quote}

\subsection{Top-Down BEV Rendering Algorithm}
\label{sec:bev-algo}

To visualize spatial configurations and egocentric camera motion in our dataset, we generate top-down bird's-eye view (BEV) maps using RGB-D and instance segmentation data. The algorithm reconstructs a 3D point cloud from RGB-D frames using camera intrinsics and extrinsics, then projects this cloud to a global top-down map. Semantic instance regions are outlined, and camera poses are rendered as arrows and trajectories.

\begin{lstlisting}[language=Python, caption={Core logic of BEV rendering using RGB-D data and camera poses.}, label={lst:bev_code}, basicstyle=\ttfamily\small, breaklines=true]
for frame in frames:
    depth = load_depth(frame)
    rgb = load_rgb(frame)
    instance_map = load_instance_seg(frame)

    rays = compute_rays(intrinsics)
    points_cam = depth * rays
    points_world = transform_to_world(points_cam, extrinsics)

    instance_ids = instance_map[valid_pixels]
    color_overlay = assign_colors(instance_ids)

    # Accumulate points for rendering
    point_cloud.append(points_world)
    colors.append(color_overlay)

# Project to 2D grid
topdown_map = render_topdown(point_cloud, colors)

# Overlay camera trajectory
plot_trajectory(poses, topdown_map)
\end{lstlisting}

\subsection{Example BEV Visualizations}
\label{sec:bev-examples}

\begin{figure*}[t]
\centering

% Left: Overlay
\begin{subfigure}[t]{0.48\textwidth}
    \centering
    \includegraphics[width=\textwidth]{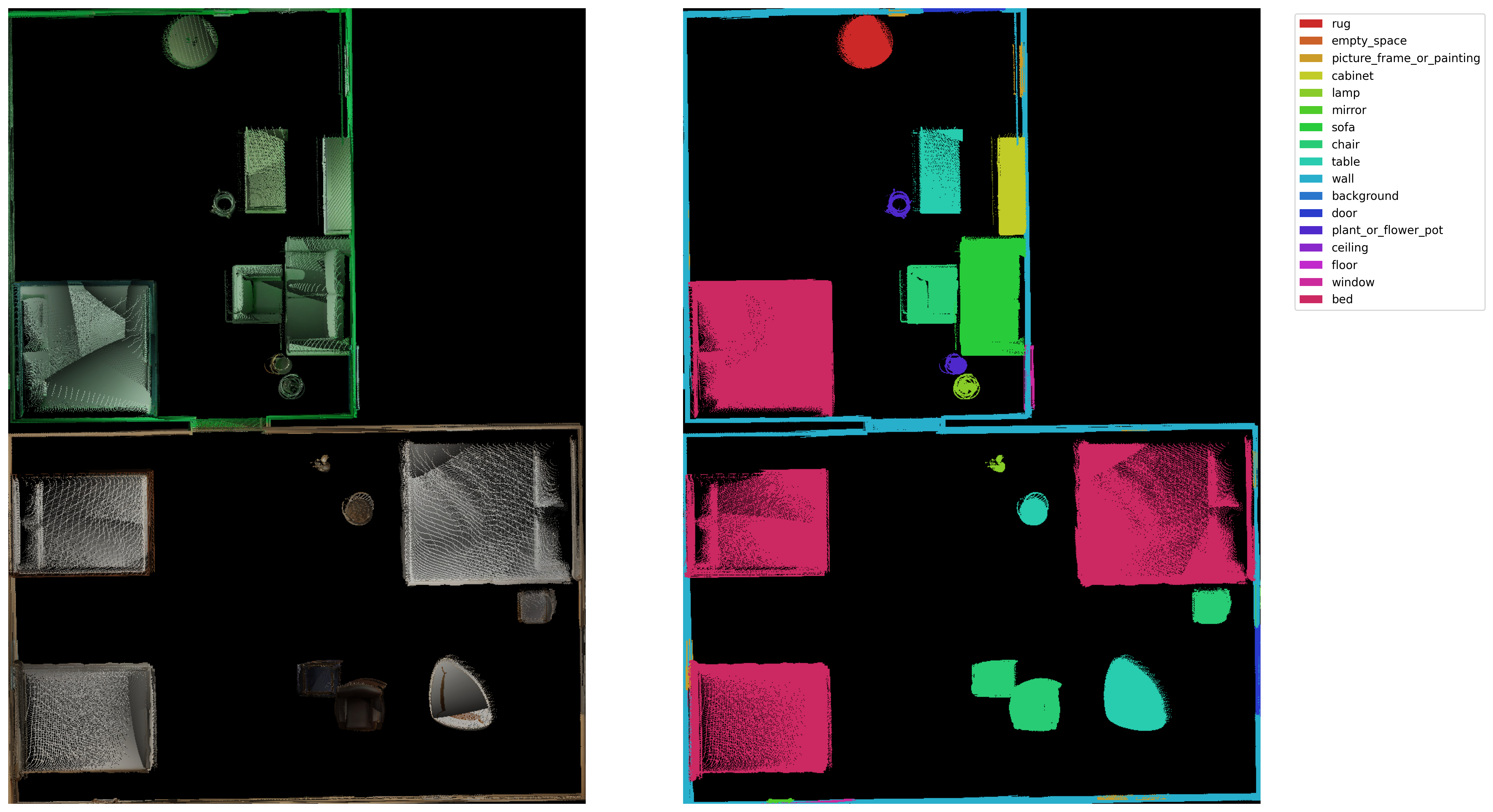}
    \caption{Top-down RGB map with object instance overlays.}
    \label{fig:bev_overlay}
\end{subfigure}
\hfill

% Right: Two side-by-side trajectory views
\begin{subfigure}[t]{0.48\textwidth}
    \centering
    \begin{subfigure}[t]{0.48\textwidth}
        \centering
        \includegraphics[width=\textwidth]{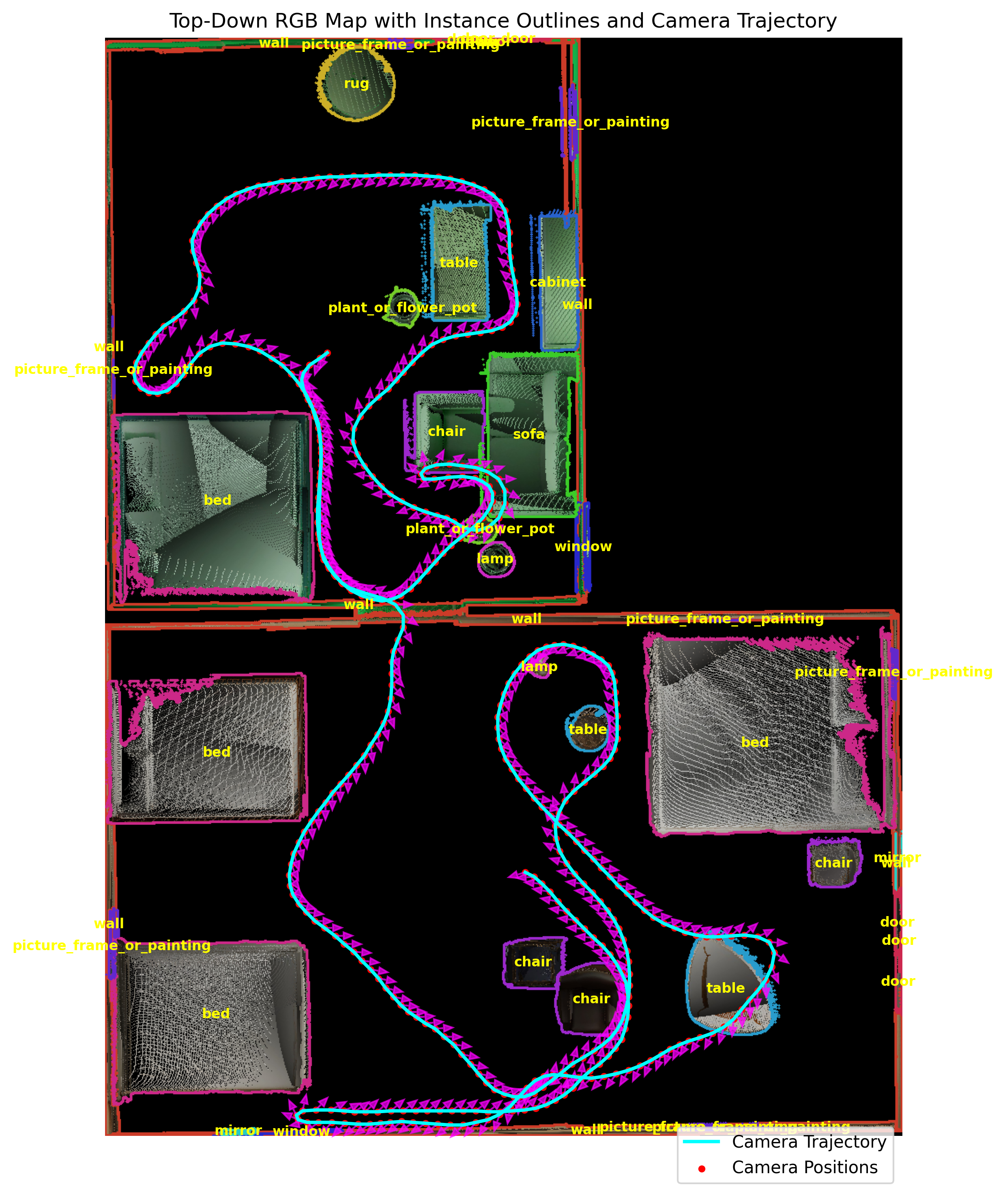}
        \caption*{Full trajectory}
    \end{subfigure}
    \hfill
    \begin{subfigure}[t]{0.48\textwidth}
        \centering
        \includegraphics[width=\textwidth]{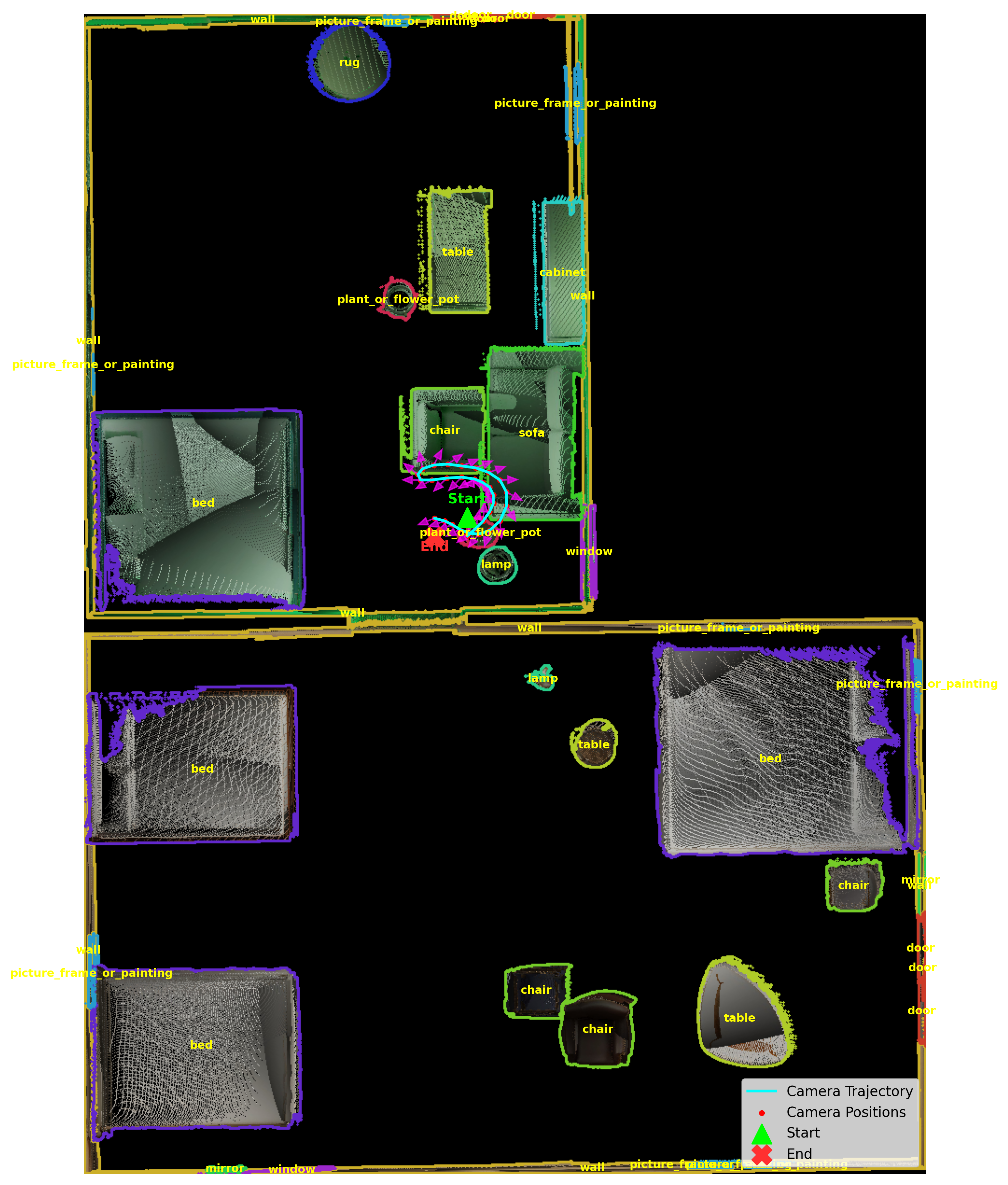}
        \caption*{Sub-trajectory (A $\rightarrow$ B)}
    \end{subfigure}
    \caption{Egocentric camera path visualizations. Left: full trajectory. Right: question-specific sub-trajectory.}
    \label{fig:bev_trajectories}
\end{subfigure}

\caption{
Top-down BEV maps for a scene in \textsc{Disjoint-3DQA}. (a) Instance-wise spatial layout reconstructed from RGB-D frames. (b) Egocentric camera trajectories showing both global and question-specific paths.
}
\label{fig:bev_examples}
\end{figure*}

Figure~\ref{fig:bev_examples} shows three BEV examples from our dataset. Each map includes semantic instance regions (outlined by color), camera trajectory (cyan line), and start/end markers. These visualizations help disambiguate spatial relations across distant or occluded frames.

\subsection{Crowdsourcing Protocol}
\label{sec:crowdsourcing}

We use Amazon Mechanical Turk (AMT) to collect human performance baselines for \datasetname{}. For each evaluation type, every sample is independently annotated by three distinct workers to ensure reliability and diversity of responses. To ensure high-quality responses, we restrict access to workers with a HIT approval rate of at least 95\% and more than 5{,}000 approved HITs. Workers are compensated at a rate of \$9.99 for answering 15 questions, estimated based on pilot timing studies. Each annotation HIT includes clear instructions and example completions. 

\begin{figure*}[ht]
    \centering
    \includegraphics[width=0.5\linewidth]{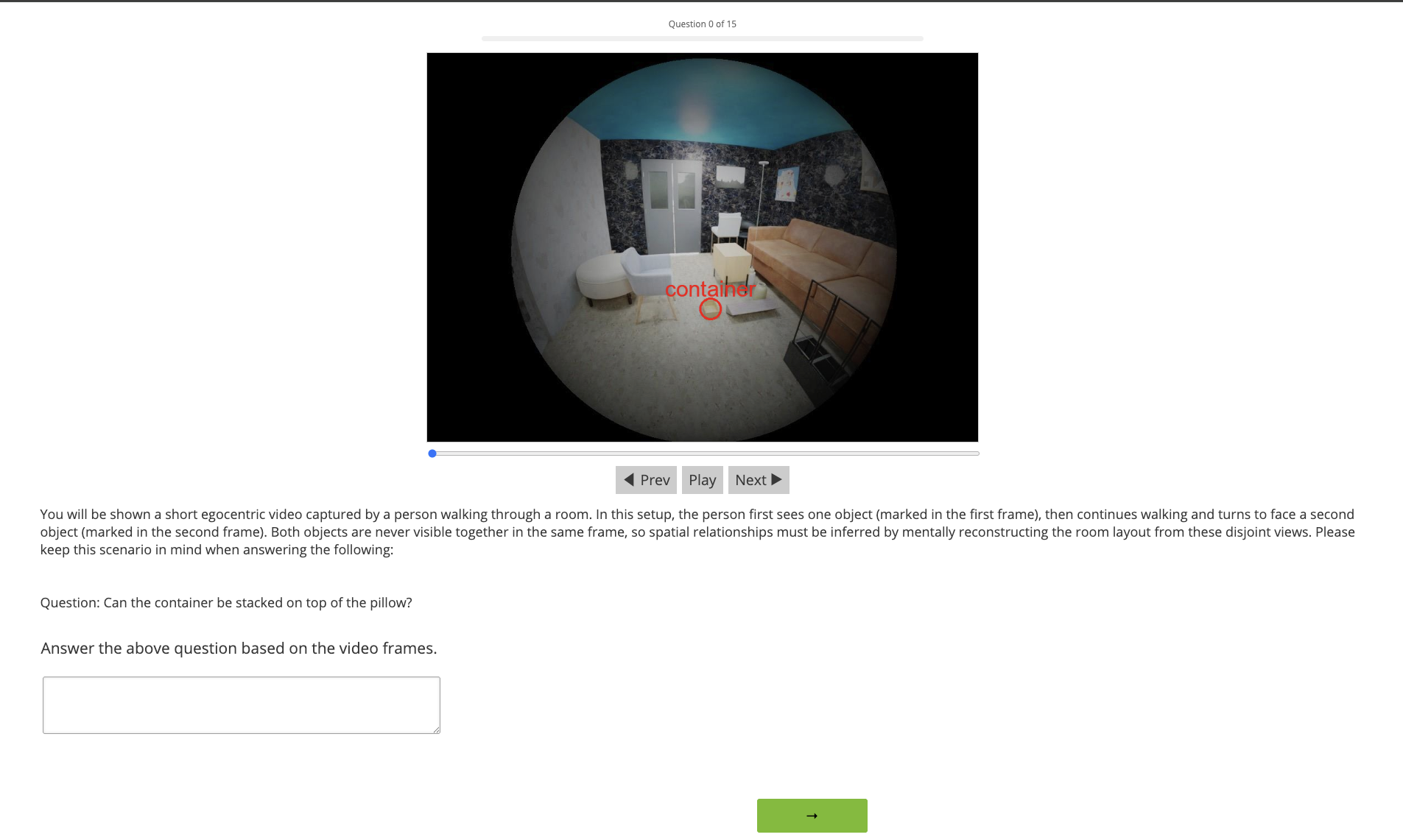}
    \caption{Template used for human evaluation}
    \label{fig:template}
\end{figure*}

We report the average response accuracy across the three annotations per example. The template used for obtaining answers to the questions is provided in Figure~\ref{fig:template} and template used for validatating \datasetname follows a similar template, with YES/NO questions about the validity of the questions, answers and bounding boxes.

\subsection{LLM-Match Prompt}
\label{sec:llm-match}

We use the following prompt to obtain semantic similarity scores for model-generated answers, following the LLM-Match protocol introduced in OpenEQA \cite{majumdar2024openeqa}. A large language model is prompted to rate the model's prediction on a scale from 1 to 5, based on its agreement with the reference and acceptable alternative answers:

\begin{quote}
You are an AI assistant who will help me to evaluate the response given the question, the correct answer, and extra answers that are also correct.
To mark a response, you should output a single integer between 1 and 5 (including 1, 5).
5 means that the response perfectly matches the answer or any of the extra answers.
1 means that the response is completely different from the answer and all of the extra answers.

\textbf{Example 1:}
Question: Is it overcast?
Answer: no
Extra Answers: \texttt{["doesn't look like it", "no", "it's sunny"]}
Response: yes
Your mark: 1

\textbf{Example 2:}
Question: Who is standing at the table?
Answer: woman
Extra Answers: \texttt{["a woman", "a lady", "woman"]}
Response: Jessica
Your mark: 3

\textbf{Example 3:}
Question: Are there drapes to the right of the bed?
Answer: yes
Extra Answers: \texttt{["yes, there are drapes", "yeah", "the drapes are to the right of the king bed"]}
Response: yes
Your mark: 5

\textbf{Your Turn:}
Question: {question}
Answer: {answer}
\end{quote}

\subsection{QA Prompt}
\label{sec:prompts}
We provide GPT-4o with a series of egocentric frames and a natural language question. Below is the standardized prompt format used during inference.

\vspace{0.5em}
\noindent\textbf{Unmarked Objects.}
\begin{quote}
You are a helpful assistant trained to answer spatial and visual questions based on egocentric video frames.
Here are the Egocentric frames:
Here is Frame 1  
\texttt{[Image 1]}

Here is Frame 2  
\texttt{[Image 2]}

...

Here is Frame N  
\texttt{[Image N]}

\texttt{[Question text]}

Please respond in the following format:

\textbf{Reason:} [Brief justification for your answer, 15--20 words] \\
\textbf{Answer:} [Concise answer, max 15--20 words]
\end{quote}
\vspace{0.5em}\
\noindent\textbf{Unmarked Objects.}
\begin{quote}
You are a helpful assistant trained to answer spatial and visual questions based on egocentric video frames.

Here are the Egocentric frames with highlighted objects in the first and last frame. The objects relevant to the question highlighted with a red hollow circle.:

Here is Frame 1 (object \texttt{A} and \texttt{B} marked)  
\texttt{[Marked Image 1]}

...

Here is Frame N (object \texttt{A} and \texttt{B} marked)  
\texttt{[Marked Image N]}

\texttt{[Question text]}

Please respond in the following format:

\textbf{Reason:} [Brief justification for your answer, 15--20 words] \\
\textbf{Answer:} [Concise answer, max 15--20 words]
\end{quote}

\subsubsection{Trajectory-Aware Prompt Format}
\label{sec:traj-prompt}

For questions where camera pose information is available, we include 3D world coordinates per frame in the prompt. The system receives a list of egocentric frames along with their estimated $(x, y, z)$ positions to help reason about spatial layout.

\vspace{0.5em}
\noindent\textbf{Example Prompt Structure:}

\begin{quote}\small
You are a spatial reasoning assistant. Here is a sequence of egocentric frames. Use the visual evidence and the associated 3D camera positions to answer the spatial question.

Frame 1 with camera's 3D position (in meters): (2.53, -1.92, 1.50)  
[Frame image]

Frame 2 with camera's 3D position (in meters): (3.18, -1.21, 1.48)  
[Frame image]

\textbf{Question:} Is the bookshelf to the left of the couch?

Please respond in the following format:  
\textbf{Reason:} [15–20 word justification]  
\textbf{Answer:} [short answer, max 20 words]
\end{quote}

\subsubsection{Top-Down BEV Prompt Format}
\label{sec:bev-prompt}

For certain questions, we supplement the egocentric video with a top-down reconstruction of the scene showing object instances and the camera trajectory.

\vspace{0.5em}
\noindent\textbf{Example Prompt Structure:}

\begin{quote}\small
The image shows a top-down view of a 3D scene reconstructed from an egocentric video.  
The magenta arrows represent the camera’s trajectory over time, based on frames relevant to the current question.  
The green triangle marks the starting camera position, and the blue X marks the ending position.

You need to focus on the part of the scene where the trajectory is marked to answer the question.

\textbf{Question:} Is the lamp behind the armchair?

Please respond in the following format:  
\textbf{Reason:} [15–20 word justification]  
\textbf{Answer:} [short answer, max 20 words]
\end{quote}

\subsection{World-to-Camera Transform \(T_B\)}  
\label{appx:camera}

Let \(c_A, c_B \in \mathbb{R}^3\) denote the world-frame centers of objects \(A\) and \(B\). We define a camera coordinate frame using the image where \(B\) is visible, placing its origin at \(c_B\) while preserving orientation. The resulting rigid-body transform is:
\begin{equation}
\begin{aligned}
T_B &=
\begin{bmatrix}
R_B & -R_B c_B \\
\mathbf{0}^\top & 1
\end{bmatrix}, \\
R_B &\in \mathrm{SO}(3), \quad T_B \in SE(3).
\end{aligned}
\label{eq:TB-def}
\end{equation} 

This transform maps a world point \(x_\text{w} \in \mathbb{R}^3\) to coordinates in the camera-\(B\) frame:
\begin{equation}
x_\text{cam} = T_B x_\text{w}, \quad x_\text{cam} \in \mathbb{R}^3.
\label{eq:TB-map}
\end{equation}

By construction, \(T_B\) maps \(c_B\) to the origin:
\begin{equation}
T_B c_B = \mathbf{0}, \quad
T_B
\begin{bmatrix}
c_B \\ 1
\end{bmatrix}
=
\begin{bmatrix}
\mathbf{0} \\ 1
\end{bmatrix}.
\label{eq:TB-zero}
\end{equation}

Applying \(T_B\) to \(c_A\) gives the coordinates of object \(A\) in the camera-\(B\) frame:
\begin{equation}
\tilde{c}_A = T_B c_A = R_B c_A - R_B c_B.
\label{eq:TB-apply}
\end{equation}

% \subsection{World-to-Camera Transform $T_{B}$}

% \label{appx:camera}

% Let \(c_{A},c_{B}\!\in\!\mathbb{R}^{3}\) denote the world–frame centres of
% objects \(A\) and \(B\).
% We attach a camera coordinate frame to the image in which \(B\) is
% visible and \emph{translate} its origin to \(c_{B}\) (the camera
% orientation is left unchanged).
% The resulting extrinsic matrix is the rigid‐body transform
% %
% \resizebox{!}{\columnwidth}{}
% {
% \begin{equation}
% T_{B} \;=\;
% \begin{bmatrix}
% R_{B} & t_{B}\\[2pt]
% \mathbf 0^{\!\top} & 1
% \end{bmatrix},
% \qquad
% R_{B}\in\mathrm{SO}(3),\;
% t_{B}=-R_{B}c_{B},
% \label{eq:TB-def}
% \end{equation}
% %
% so \(T_{B}\!\in\!SE(3)\).

% The transform maps world–frame points
% \(x_{\text{w}}\in\mathbb{R}^{3}\) (or homogeneous
% \([\;x_{\text{w}}\;1]^{\!\top}\in\mathbb{R}^{4}\))
% to camera-\(B\) coordinates:
% %
% \begin{equation}
% x_{\text{cam}}
% \;=\;
% T_{B}\,x_{\text{w}},
% \qquad
% x_{\text{cam}}\in\mathbb{R}^{3}.
% \label{eq:TB-map}
% \end{equation}

% Because \(t_{B}\) is chosen as in~\eqref{eq:TB-def},
% %
% \begin{equation}
% T_{B}c_{B}= \mathbf 0,
% \qquad
% T_{B}
% \begin{bmatrix}
% c_{B}\\[2pt]1
% \end{bmatrix}
% =
% \begin{bmatrix}
% \mathbf 0\\[2pt]1
% \end{bmatrix},
% \label{eq:TB-zero}
% \end{equation}
% %
% so object \(B\) is located at the origin of the camera-\(B\) frame.

% Applying \(T_{B}\) to \(c_{A}\) yields the centre of \(A\) expressed in
% camera-\(B\) coordinates:
% %
% \begin{equation}
% \tilde c_{A}
% \;=\;
% T_{B}\,c_{A}
% \;=\;
% R_{B}c_{A}+t_{B}.
% \label{eq:TB-apply}
% \end{equation}
% }

\end{document}